%% file: main.tex
\documentclass[conference]{IEEEtran}

\IEEEoverridecommandlockouts

\usepackage{cite}
\usepackage{amsmath,amssymb,amsfonts}
\usepackage{algorithmic}
\usepackage{algorithm}
\usepackage{graphicx}
\usepackage{textcomp}
\usepackage{xcolor}
\usepackage{url}
\usepackage{booktabs}

\def\BibTeX{{\rm B\kern-.05em{\sc i\kern-.025em b}\kern-.08em
T\kern-.1667em\lower.7ex\hbox{E}\kern-.125emX}}

\begin{document}

\title{Adaptive Group-Based Counterfactual Explanations for Time-Series Rehabilitation Data\\
\thanks{This work was supported in part by NWO LoaD-project. NWA.1389.20.009}
}

\author{
\IEEEauthorblockN{1\textsuperscript{st} Emmanuel C. Chukwu}
\IEEEauthorblockA{
\textit{Eindhoven University of Technology} \\
Eindhoven, the Netherlands \\
e.c.chukwu@tue.nl}
\and
\IEEEauthorblockN{2\textsuperscript{nd} Rianne M. Schouten}
\IEEEauthorblockA{
\textit{Eindhoven University of Technology} \\
Eindhoven, the Netherlands \\
r.m.schouten@tue.nl}
\and
\IEEEauthorblockN{3\textsuperscript{rd} Monique Tabak}
\IEEEauthorblockA{
\textit{University of Twente} \\
Enschede, the Netherlands \\
m.tabak@utwente.nl}
\and
\IEEEauthorblockN{4\textsuperscript{th} Mykola Pechenizkiy}
\IEEEauthorblockA{
\textit{Eindhoven University of Technology} \\
Eindhoven, the Netherlands \\
m.pechenizkiy@tue.nl}
}

\maketitle
\begin{abstract}
Counterfactual explanations (CEs) for multivariate time-series classifiers are often difficult to interpret in domains where experts reason in terms of semantic feature groups rather than individual channels. In rehabilitation movement analysis with multi-sensor inertial measurement units (IMUs), clinicians interpret motion through muscle-group and joint-segment abstractions; yet, most existing counterfactual methods operate at the channel level, producing scattered and biomechanically incoherent explanations. We propose a two-stage framework for group-based counterfactual generation in high-dimensional IMU data. We first show that Shapley-Adaptive (SA) group ranking preserves counterfactual validity but fails to enforce group-level sparsity, motivating the need for explicit group selection. We then introduce Learnable Gate (LG) methods, which incorporate trainable per-group relevance gates jointly optimized with perturbation masks. Experiments on the KneE-PAD rehabilitation dataset demonstrate that LG substantially improves modality-group sparsity compared to the channel-level M-CELS baseline while maintaining or improving validity, temporal smoothness, and generation efficiency. Exercise-specific analyses further show that group-structured counterfactuals yield concise, muscle-level corrective guidance aligned with clinical reasoning. Overall, the proposed framework enhances interpretability without sacrificing counterfactual quality, enabling more actionable explanations for rehabilitation movement analysis.
\end{abstract}


\section{Introduction}
\label{sec:introduction}

\subsection{Motivation}

Recent advances in machine learning have improved analysis of complex movement patterns for clinical rehabilitation and injury assessment\cite{Mahmoud2025,Nicora2025}. For clinical use, however, models must offer explanations that are interpretable, grounded in expert domain concepts, and actionable, providing concrete guidance for intervention. In rehabilitation, this requires explanations aligned with biomechanical principles and anatomical structures familiar to patients and physical therapists, rather than abstract feature indices\cite{Rosa2024}. Standard attribution methods such as SHAP, when applied to IMU time series, explain predictions at the feature or channel level, offering limited insight into biomechanically meaningful units like muscle groups or joint coordination patterns\cite{lundberg2017unified}.

CEs, which identify \emph{sufficient changes} to flip a prediction, are particularly promising for high-stakes domains like rehabilitation, where ``what-if'' scenarios can guide personalized interventions\cite{wachter2017counterfactual}. Yet for high-dimensional multi-IMU time series, existing approaches typically operate on individual channels or time steps\cite{li2024mcels,hollig_tsevo_2022,ates2021comte}. This causes several issues: (i) combinatorial explosion, as the search space scales with channels times time steps, often producing dense perturbations that alter many scattered features;\cite{ates2021comte,delaney2021instance} (ii) violation of physical and clinical constraints, since channel-wise edits can yield biomechanically implausible sequences, such as isolated axis changes that ignore segment-level coupling and joint kinematics;\cite{zhu2025imu} (iii) temporal fragmentation, where dispersed edits over isolated time steps create ``temporal confetti'' rather than coherent intervention windows;\cite{delaney2021instance} and (iv) flat explanations that lack a hierarchy from system-level concepts down to specific sensors\cite{Rosa2024}.

These limitations are exacerbated in multi-IMU settings: 48 channels are structured as 6-channel accelerometer–gyroscope units per anatomical site, while clinicians reason in terms of segments, joints, and synergies\cite{Rosa2024,Porciuncula2018}. Clinical IMU studies emphasize segment/joint-level assessment\cite{Routhier2020,Kasnesis2025} and anatomically valid sensor placement and calibration\cite{niswander2020placement,bonfiglio2024calibration}. Channel-level CFEs may be sparse yet biomechanically incoherent, scattering changes across sensors. Group-based CFEs operate on semantic sensor groups (e.g., ``reduce right hamstring activity during squat descent''), yielding minimal, plausible, temporally coherent interventions. While tabular CFEs often incorporate feasibility, causal, or correlation constraints \cite{karimi2021algorithmic,mahajan2019preserving,pawelczyk2020learning}, time-series methods rarely encode sensor grouping or segment-level abstractions, limiting clinical usefulness in higher dimensional space.

\subsection{Research Objective}
This work studies whether adaptive, group-based counterfactuals improve interpretability and reliability in high-dimensional IMU time-series classification. We ask: \emph{can dynamic group-level optimization yield more interpretable and reliable explanations than channel-level methods?} We hypothesize that domain-aligned grouping improves interpretability by matching clinical structure\cite{sergeev2025data}, and that dynamic selection with multi-objective optimization can balance validity, sparsity, and plausibility\cite{dandl2020multi,mothilal2020explaining}. We evaluate on KneE-PAD across squat, knee extension, and gait, comparing against a strong channel-level baseline\cite{Kasnesis2025}.

\subsection{Contributions}

This paper makes three contributions: (i) an adaptive multi-objective framework (Adaptive-MO) that combines Shapley-based group ranking with dynamic sensor group selection to generate structured CEs for high-dimensional time series; (ii) a Learnable Gate (LG) mechanism with trainable per-group relevance parameters, jointly optimized with perturbation masks, to automatically select a sparse, interpretable, and clinically meaningful set of sensor groups; and (iii) validation on the KneE-PAD dataset, showing, via systematic and expert-aligned evaluation, that group-structured counterfactuals produce concise corrective feedback across multiple rehabilitation exercises through structured, sparser, and interpretable changes. The project repository is available at \url{https://github.com/Healthpy/knee-rehab-cfe}.

\section{Related Work}
\label{sec:related}

\subsection{Counterfactual Explanations for Time Series}

A broad range of CE methods have been proposed for time-series classification, including instance-based, shapelet-guided, saliency- and attention-driven, gradient-based, latent generative, and evolutionary approaches\cite{ates2021comte,delaney2021instance,li_sg-cf_2022,huang_shapelet-based_2024,filali2022mining,li2024mcels,li_attention_2023,wang_learning_2021,van_looveren_conditional_2021,hollig_tsevo_2022}. Instance-focused methods generate counterfactuals using nearest unlike neighbors or distractor examples, while shapelet-based approaches exploit discriminative subsequences to guide structured modifications \cite{ates2021comte,delaney2021instance,li_sg-cf_2022,huang_shapelet-based_2024}. Saliency and attention methods perturb influential time intervals, whereas gradient-based and latent generative techniques optimize smooth and plausible changes in input or latent space\cite{filali2022mining,li2024mcels,li_attention_2023,wang_learning_2021,van_looveren_conditional_2021}. Evolutionary methods further explore the counterfactual space through multi-objective optimization or sequential decision-making\cite{hollig_tsevo_2022}.

Most existing techniques focus on individual features, time steps, or short subsequences, limiting their ability to capture higher-level structures such as sensor- or modality-level groupings. Consequently, many CEs lose clinical meaning in high-dimensional multivariate settings and face the combinatorial, stability, plausibility, temporal, and hierarchical issues outlined in Section~\ref{sec:introduction}. This motivates counterfactual frameworks that explicitly model structured dependencies and apply domain-informed constraints to yield valid, interpretable explanations in real multichannel scenarios.

\subsection{Movement Analysis, IMUs, and Rehabilitation}

IMUs are increasingly used to quantitatively assess human movement in clinical practice\cite{Routhier2020,dalfarra2025,Porciuncula2018}. Traditional IMU analyses often rely on handcrafted features such as angular velocity, joint angles, or gait parameters,\cite{cutti2010imu} whereas deep learning on raw sequences improves accuracy while reducing interpretability\cite{xu2021,li2022calib,li2022interpretable}. Clinical interpretability requires mapping sensor readings to biomechanically meaningful constructs (e.g., limb orientation, angular velocity patterns, compensatory asymmetries)\cite{zhu2025imu}. Recent work emphasizes sensor placement and calibration as critical steps for obtaining anatomically valid joint kinematics from IMUs\cite{niswander2020placement,bonfiglio2024calibration}, and qualitative studies highlight that clinicians and patients prefer summaries at the joint or segment level over raw signals\cite{Routhier2020}.

The KneE-PAD dataset demonstrates the feasibility of large-scale, sensor-based assessment of knee rehabilitation exercises with 8 IMU sensors placed on key lower-limb muscle groups\cite{Kasnesis2025}. Explanatory methods should therefore preserve relationships between accelerometer and gyroscope components within each sensor unit and between sensors on anatomically linked segments, exactly as encoded in our grouping strategy.

\subsection{Group-Level Feature Selection and Attribution}

Feature grouping and structured sparsity have been widely studied in statistics and machine learning, including Group LASSO\cite{yuan2006model}, structured sparsity-inducing norms\cite{jenatton2011structured}, and hierarchical importance measures\cite{fisher2019all}. Shapley values provide principled attribution and can be extended to group-level importance\cite{lundberg2017unified,sundararajan2020many,covert2021explaining}. In CEs more broadly, evaluation rubrics emphasize desiderata such as validity, sparsity, proximity, plausibility, robustness, and diversity\cite{verma2024}. However, the use of group-based importance and structured sparsity in time-series CEs, particularly for sensor-based human movement analysis, remains largely unexplored. Our work leverages Shapley-based ranking at the group level, combined with adaptive group gating, to create a structured and physiologically meaningful counterfactual search space.

\subsection{M-CELS and Sequence-Based Counterfactual Baselines}

M-CELS is a recent state-of-the-art counterfactual method for multivariate time-series data.\cite{li2024mcels} M-CELS learns a continuous temporal saliency mask and constructs counterfactuals by blending the input with a nearest-unlike-neighbor guide example. Its optimization balances three objectives: (i) achieving the target prediction, (ii) minimizing the magnitude of the mask (sparsity), and (iii) enforcing temporal smoothness. However, M-CELS treats all channels independently, without any notion of sensor structure or modality-specific grouping. This results in modifications across many channels and therefore offers limited interpretability in domains where channels can naturally be grouped into anatomical or physical units. Our framework uses M-CELS as a strong channel-level baseline and extends this paradigm with group-based ranking and optimization.

\section{Problem Formulation and Background}
\label{sec:background}

\subsection{Time-Series Classification Setup}

Let \(X \in \mathbb{R}^{C \times T}\) denote a multivariate time series with \(C\) channels and \(T\) time steps. In our IMU setup, \(C = 48\) channels from 8 sensors, each providing 3-axis accelerometer and 3-axis gyroscope signals. A trained classifier \(f: \mathbb{R}^{C \times T} \to \{0,\dots,8\}\) predicts one of nine exercise variations (three exercises times three execution types: correct and two clinically defined errors). For an instance \(X\) with predicted label \(y = f(X)\), a CE seeks a modified input \(\tilde{X}\) such that \(f(\tilde{X}) = y_{\text{target}} \neq y\), while keeping \(\tilde{X}\) close to \(X\) according to a distance measure \(d(X,\tilde{X})\) and additional constraints (e.g., sparsity, plausibility).

\subsection{Group-Based Counterfactuals}

To address the limitations of channel-wise counterfactuals, we define a set of sensor groups \(G = \{g_1,\dots,g_K\}\), where each \(g_k \subseteq \{1,\dots,C\}\) corresponds to an anatomical IMU location (e.g., right rectus femoris, left gastrocnemius). Let \(X_{g_k} \in \mathbb{R}^{|g_k| \times T}\) denote the sub-series associated with group \(g_k\). Instead of optimizing per-channel changes, we consider group-level perturbations and define a group-wise distance \(D_G(X,\tilde{X}) = \sum_{k=1}^{K} w_k \, d\big(X_{g_k}, \tilde{X}_{g_k}\big)\), where \(w_k\) are group weights. Group sparsity is measured by counting the number of groups with non-zero perturbations, \(\text{G\_Sparsity}(X,\tilde{X}) = \sum_{k=1}^{K} \mathbb{I}\!\left(\|X_{g_k} - \tilde{X}_{g_k}\|_2 > \epsilon\right)\). Our objective is to generate counterfactuals that (i) flip the prediction to \(y_{\text{target}}\), (ii) use as few sensor groups as possible, and (iii) remain close to the data manifold and biomechanically plausible.

\section{Proposed Method}
\label{sec:methodology}

Our Adaptive-MO framework consists of three main components: (1) domain-informed sensor grouping, (2) Shapley-based group ranking, and (3) learnable group gates combined with adaptive multi-objective optimization.

\subsection{Domain-Informed Grouping}

We consider two IMU channel grouping strategies: (i) a \emph{sensor-based} grouping, in which each of the eight IMU sensors forms a group containing its six channels (three-axis accelerometer and three-axis gyroscope), yielding $K=8$ groups over $C=48$ channels; and (ii) a \emph{modality-based} grouping, in which accelerometer and gyroscope signals are treated as separate groups for each muscle, yielding $K=16$ modality-specific groups. We adopt the modality-based strategy, as it provides finer granularity by explicitly distinguishing linear acceleration from rotational kinematics at the muscle level, enabling clearer biomechanical attribution of counterfactual modifications while preserving interpretability. Sensor channels are indexed sequentially by muscle and body side, with each muscle-specific sensor comprising three accelerometer channels followed by three gyroscope channels; for example, the right rectus femoris (R\_RF) corresponds to accelerometer channels 1–3 and gyroscope channels 4–6, with subsequent muscle groups following the same indexing pattern.

\subsection{Shapley-Based Group Ranking}

Given an instance \(X\) and classifier \(f\), we first estimate channel-level Shapley values \(\phi_i\) using GradientSHAP.\cite{lundberg2017unified} Channel attributions are then aggregated to sensor groups using the maximum absolute value:
\begin{equation}
\Phi_g = \max_{i \in g} \bar{|\phi_i|},
\label{eq:groupimportance}
\end{equation}
where \(\bar{|\phi_i|}\) is the mean absolute Shapley value of channel \(i\) across time steps. Using the maximum rather than the mean ensures that a single highly important channel elevates the entire group, avoiding dilution when other channels have near-zero importance.

We then select the top-\(k\) most influential groups:
\[
G_{\text{sel}} = \text{Top-}k\{\Phi_g : g \in G\}, \quad k = \left\lfloor r \cdot K \right\rfloor,
\]
where \(r \in (0,1]\) is the group selection ratio (we use \(r = 0.8\) in our experiments). Restricting optimization to \(G_{\text{sel}}\) reduces search complexity and encourages group-level sparsity.

\subsection{Counterfactual Parameterization}

Following the M-CELS paradigm\cite{li2024mcels}, we parameterize counterfactuals via a continuous modification mask \(M \in [0,1]^{C \times T}\) that blends the original input \(X\) with a guide example \(X_{\text{guide}}\) from the target class:
\begin{equation}
\tilde{X} = X \odot (1 - M) + X_{\text{guide}} \odot M,
\label{eq:blend}
\end{equation}
where \(\odot\) denotes element-wise multiplication. The guide example is chosen as the nearest unlike neighbor with label \(y_{\text{target}}\):
\[
X_{\text{guide}} = \arg\min_{X' : f(X') = y_{\text{target}}} \|X - X'\|_2.
\]

We maintain a restricted mask \(M_{\text{rest}}\) only for channels in \(G_{\text{sel}}\), which is then expanded to the full mask according to group structure and modulated by learnable group gates.

\subsection{Learnable Group Gates}

To refine which groups are used in the counterfactual, we introduce learnable gates \(\gamma_g \in [0,1]\) for \(g \in G_{\text{sel}}\) that modulate the mask per group:
\begin{equation}
M_{i,t} = \sigma(M_{\text{rest},i,t}) \cdot \gamma_g, \quad \forall i \in g,\, g \in G_{\text{sel}},
\label{eq:gating}
\end{equation}
where \(\sigma\) is the sigmoid function. Groups with \(\gamma_g \approx 0\) are effectively dropped, enabling data-driven group selection during optimization.

We regularize the gates with a combined \(\ell_1\) and binarization penalty:
\[
L_{\text{gates}} = 
\underbrace{\frac{1}{|G_{\text{sel}}|} \sum_{g \in G_{\text{sel}}} \gamma_g}_{\text{\(\ell_1\): drives unused gates to 0}} 
+
\underbrace{\frac{1}{|G_{\text{sel}}|} \sum_{g \in G_{\text{sel}}} \min(\gamma_g, 1 - \gamma_g)}_{\text{binarization: pushes gates to 0 or 1}}.
\]
The gates \(\gamma_g\) control which sensor groups affect the counterfactual. The \(\ell_1\) term encourages sparsity by penalizing all gate activations, while the binarization term penalizes intermediate values, pushing gates toward 0 or 1. Applied with a warm-up and gradual ramp-up to avoid premature pruning, this regularization converts the SHAP-selected set \(G_{\text{sel}}\) into a sparse, binary subset \(G'_{\text{sel}} \subseteq G_{\text{sel}}\), yielding interpretable explanations where each retained group has a clear role in the counterfactual.

\subsection{Multi-Objective Loss}

We combine four complementary loss terms:
\begin{equation}
L_{\text{total}} = w_1 L_{\text{target}}
+ w_2 L_{\text{sparsity}}
+ w_3 L_{\text{smooth}}
+ w_4 L_{\text{gates}},
\label{eq:multiobjective}
\end{equation}

where $L_{\text{target}} = 1 - p_{y_{\text{target}}}$ enforces that $f(\tilde{X})$ predicts the target class with high probability $p_{y_{\text{target}}}$, $L_{\text{sparsity}} = \frac{1}{CT}\lVert M\rVert_1$ encourages channel-level sparsity, $L_{\text{smooth}} = \operatorname{mean}_{t}\!\left(\lVert M_{\cdot,t+1} - M_{\cdot,t}\rVert_1^{\alpha}\right)$ with $\alpha = 3$ enforces smoothness via a higher-order total variation norm, and $L_{\text{gates}}$ denotes the gate regularization term defined above. The weights \(w_i\) are adapted during optimization via multiplicative scaling based on the current target probability \(p_{y_{\text{target}}}\). For \(w_1\), \(w_2\) and \(w_3\) we use
\[
w_1^{(t+1)} =
\begin{cases}
0.95 \cdot w_1^{(t)} & \text{if } p_{y_{\text{target}}} \ge \tau, \\
1.05 \cdot w_1^{(t)} & \text{otherwise,}
\end{cases}
\]

\[
w_{2,3}^{(t+1)} =
\begin{cases}
1.05 \cdot w_{2,3}^{(t)} & \text{if } p_{y_{\text{target}}} \ge \tau, \\
0.95 \cdot w_{2,3}^{(t)} & \text{otherwise.}
\end{cases}
\]
All weights are clamped to \([0.1, 2.0]\), with a target threshold \(\tau\). This multiplicative schedule gradually shifts emphasis from validity to sparsity and structure once a confident prediction is achieved.

\subsection{Optimization Procedure and Post Refinement}

Algorithm~\ref{alg:adaptive_mo} summarizes the procedure. It generates counterfactuals on channel groups rather than individual channels. It computes channel importance with GradientSHAP, aggregates scores into anatomically meaningful groups (line 1), and selects the top-k most influential groups to restrict the search (line 2). A nearest neighbor from the target class serves as a guide (line 3), and the method initializes a continuous perturbation mask plus learnable group gates that start mostly active but remain prunable (line 4), with weights initialized (line 5). The algorithm then iteratively blends the original input with the guide (lines 7–8) while jointly optimizing four objectives: prediction validity, sparsity, temporal smoothness, and gate regularization (line 9). An adaptive weighting scheme dynamically shifts emphasis from validity to sparsity once the target probability is reached (line 10), enabling early stopping on convergence (lines 11–13). The key innovation is the learnable group gates, which suppress irrelevant groups during optimization and produce structured counterfactuals focused on a few interpretable, meaningful sensor locations.

\begin{algorithm}[t]
\caption{Adaptive Multi-Objective Group-Based CFE}
\label{alg:adaptive_mo}
\begin{algorithmic}[1]
\REQUIRE Input \(X\), model \(f\), 
class 
\(y_{\text{target}}\), background set \(B\)
\STATE Compute channel Shapley values \(\phi_i\) and aggregate to groups \(\Phi_g\) using \eqref{eq:groupimportance}
\STATE \(G_{\text{sel}} \gets\) top-\(k\) groups by \(\Phi_g\)
\STATE \(X_{\text{guide}} \gets \arg\min_{X' \in B : f(X') = y_{\text{target}}} \|X - X'\|_2\)
\STATE Initialize \(M_{\text{rest}} \sim \mathcal{U}(0,1)\) for channels in \(G_{\text{sel}}\); gate logits \(\ell_g \gets 2.0\) so \(\gamma_g = \sigma(\ell_g) \approx 0.88\)
\STATE Initialize weights \(w_1 = 1.0\), \(w_{2\text{--}4} = 0.5\); freeze gates during warm-up
\FOR{\(t = 1\) to \(T_{\max}\)}
    \STATE Construct \(M\) using \eqref{eq:gating} and group expansion
    \STATE \(\tilde{X} \gets X \odot (1 - M) + X_{\text{guide}} \odot M\) using \eqref{eq:blend}
    \STATE Compute losses \(L_{\text{target}}, L_{\text{sparsity}}, L_{\text{smooth}}, L_{\text{gates}}\)
    \STATE Update weights \(w_i\) based on \(p_{y_{\text{target}}}\) and \(\tau\)
    \STATE \(L_{\text{total}} \gets \sum_i w_i L_i\)
    \STATE Take a gradient step on \(M_{\text{rest}}\) and \(\gamma_g\) 
    \IF{\(p_{y_{\text{target}}} \ge \tau\) and convergence criterion satisfied}
        \STATE \textbf{break}
    \ENDIF
\ENDFOR
\RETURN Counterfactual \(\tilde{X}\), mask \(M\), selected groups \(G'_{\text{sel}}\)
\end{algorithmic}
\end{algorithm}


After convergence of the learnable-gate optimization, we apply a lightweight post-processing step to extract the sparsest valid counterfactual. Because joint optimization over the continuous mask \(M\) and gate logits \(\ell_g\) balances multiple objectives, it typically yields a valid but not maximally sparse solution once \(p_{y_{\text{target}}} \ge \tau\), with little incentive to further reduce mask support. We therefore smooth the learned mask temporally and perform a small grid search over binarization thresholds and optional temporal blending. All candidate masks are validated against the target constraint, and the sparsest valid one is chosen, breaking ties by minimal temporal variation. This inexpensive refinement runs once after convergence.

\section{Experimental Setup}
\label{sec:experiments}

\subsection{Dataset and Exercise Protocol}

We evaluate our framework on the KneE-PAD dataset,\cite{Kasnesis2025} which includes three lower-limb rehabilitation exercises: squat, seated knee extension, and gait, performed by 31 patients with knee pathologies. For each exercise, physiotherapists defined one correct execution and two relevant incorrect variations. For squats, the incorrect variations are weight shifting to the healthy leg during descent (SquatWT) and placing the injured leg forward (SquatFL). For seated knee extensions, the incorrect forms include incomplete extension, limiting the range of motion (ExtNF), and lateral abduction of the injured leg from the intended plane (ExtLL). For gait, incorrect variations consist of failure to fully extend the injured leg during stance (GaitNF) and lateral abduction altering its trajectory (GaitHA). Participants wore eight IMU sensors placed bilaterally on key lower-limb muscle groups: rectus femoris (RF), hamstrings (HAM), tibialis anterior (TA), and gastrocnemius (GAS) on each leg. Each IMU recorded tri-axial accelerometer and gyroscope data, yielding 48 channels. Synchronized sEMG was also recorded but is not used directly in this study.

\subsection{Preprocessing and Data Splits}

Each recording was segmented into fixed-length windows covering a full exercise repetition. To ensure generalization and prevent subject leakage, we use a subject-disjoint split: subjects are divided into separate training, validation, and test sets. A greedy bin-packing heuristic assigns subjects to approximate a 70/15/15 train/val/test ratio by sample count, with no subject shared across splits. Data normalization (z-score) is computed on the training set only and applied to all splits, evaluating generalization to unseen subjects.

\subsection{Classification Model}

Our classifier is a fully convolutional network (FCN)\cite{wang2017time} with three 1D convolutional blocks (48\(\to\)128\(\to\)256\(\to\)128 channels) using kernel sizes 8, 5, and 3, each followed by batch normalization and ReLU activations. A global average pooling layer aggregates temporal features, followed by dropout (\(p = 0.2\)) and a fully connected layer for nine-class classification (3 exercises \(\times\) 3 execution types). The model uses IMU-only input (48 channels from 8 bilateral sensors). We train with AdamW (learning rate \(10^{-3}\), weight decay \(10^{-5}\)), cross-entropy loss, batch size 32, and up to 50 epochs with ReduceLROnPlateau (factor 0.5, patience 5) on the subject-disjoint validation set. Validation-based learning rate reduction and checkpointing on best validation accuracy provide implicit early stopping.

\subsection{Methods}

The classifier \(f:\mathbb{R}^{C\times T} \rightarrow \{0,\ldots,8\}\) predicts nine variants across three rehabilitation exercises: correct executions \(\{0,3,6\}\) (squat, seated knee extension, gait) and exercise-specific error classes \(\{1,2\}\), \(\{4,5\}\), and \(\{7,8\}\). Counterfactual generation seeks a minimally modified erroneous input \(X\) such that \(f(\tilde{X}) = y_{\text{target}} \in \{0,3,6\}\) for the same exercise. This error-to-correct setting yields meaningful guidance by identifying minimal, group-sparse sensor changes needed to correct movement execution. We compare eight method configurations in three families:

\paragraph{Baseline: M-CELS.} M-CELS is a channel-level counterfactual method using multi-objective optimization (validity, sparsity, temporal smoothness)\cite{li2024mcels}. It treats all 48 channels independently and serves as our channel-level baseline.

\paragraph{Shapley-Adaptive (SA) variants.} SA ranks feature groups with GradientSHAP and perturbs only top-ranked groups. We test four variants: \textbf{SA (SHAP pruned)} (keeps top \(r = 0.8\) SHAP-ranked groups; uses group-aware loss), \textbf{SA}-no group loss (removes group sparsity penalty), \textbf{SA}-no adaptive (removes adaptive weight scheduling), and \textbf{SA}-no SHAP, all groups (no SHAP ranking; perturbs all groups equally).

\paragraph{Learnable Gate (LG) variants.} LG adds trainable sigmoid group gates \(\gamma_g\) (Section~\ref{sec:methodology}) to mask-based counterfactuals for adaptive group relevance. We test three variants: \textbf{LG-scratch} (uniform gate initialization without SHAP; prunes gates via \(\tau_{\text{adapt}} = \mathrm{mean}(\gamma_g) - \mathrm{std}(\gamma_g)\)), \textbf{LG-SHAP pruned} (SHAP-initialized gates on top \(r = 0.8\) groups, then adaptively pruned), and \textbf{LG-fixed prune} (SHAP preselection with learnable gates, pruned at fixed \(\tau = 0.3\)).

\subsection{Evaluation Metrics}

We evaluate CEs using six metrics inspired by established counterfactual XAI evaluation rubrics~\cite{verma2024}: \textbf{validity}, defined as the fraction of instances for which at least one counterfactual satisfies $f(\tilde{X}) = y_{\text{target}}$ with $p_{y_{\text{target}}} \ge 0.8$; \textbf{sparsity}, measured as the fraction of unchanged entries in $X$; \textbf{channel sparsity}, quantified by the number of channels entirely unchanged over time; \textbf{group sparsity}, defined as the number of unmodified modality groups; \textbf{proximity}, measured by the $L_2$ distance between $X$ and $\tilde{X}$; and \textbf{efficiency and plausibility}, assessed via wall-clock generation time, optimization steps to convergence, and the temporal gradient as a measure of smoothness.


To ensure a fair comparison, we standardize the core optimization settings for Adaptive-MO and M-CELS (Table~\ref{tab:hyperparams}).

\begin{table}[t]
\centering
\caption{Hyperparameters for SA, LG, and M-CELS. Unless stated otherwise, learning rate = 0.01, max iterations = 5000, mask threshold = 0.5, LR decay \((\gamma)\) = 0.9991, Optimizer = Adam}
\label{tab:hyperparams}
\begin{tabular}{lccc}
\toprule
Parameter & SA & LG & M-CELS \\
\midrule
Target threshold \((\tau)\)   & 0.72  & 0.75  & 0.70 \\
Group selection ratio \((r)\) & 0.8   & 0.8   & --   \\
Gate LR multiplier            & --    & 1.0   & --   \\
Gate warm-up iters            & --    & 150   & --   \\
\(w_{\text{target}}\) (init.) & 1.0   & 1.0   & 0.7  \\
\(w_{\text{sparsity}}\) (init.) & 0.55 & 0.75 & 0.5  \\
\(w_{\text{smooth}}\) (init.)   & 0.50 & 0.65 & 0.6  \\
\(w_{\text{gates}}\) (init.)    & --   & 0.90 & --   \\
\bottomrule
\end{tabular}
\end{table}

\section{Results}
\label{sec:results}

All results use the single-seed, subject-disjoint protocol (seed 42, \(n = 150\) test samples) described in Section~\ref{sec:experiments}.

\subsection{Shapley-Adaptive Group Ratio Analysis}

\begin{figure}[b]
\centering
\includegraphics[width=0.75\linewidth]{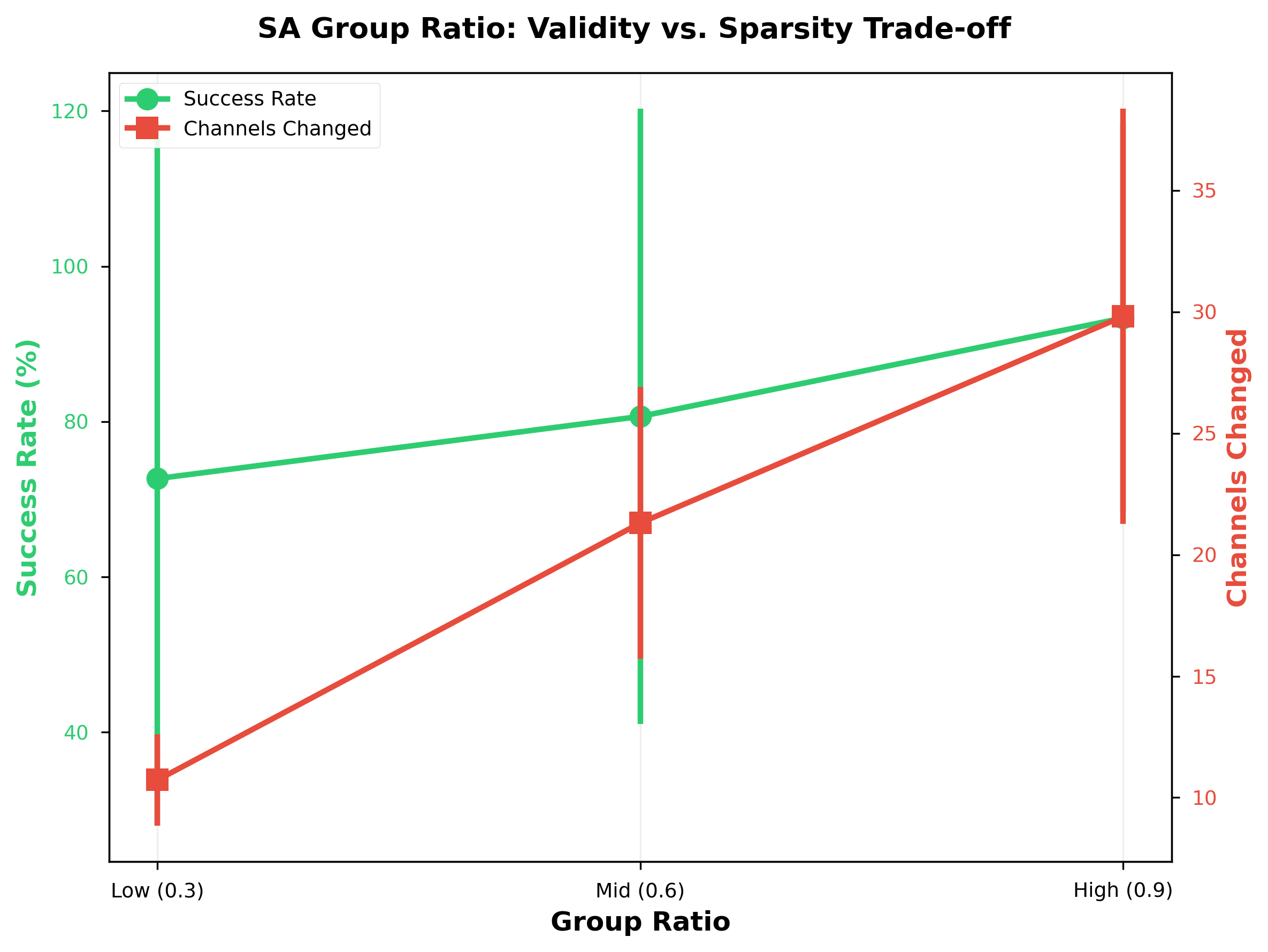}
\caption{SA group ratio impact on validity vs.\ sparsity (\(n = 150\)). Higher ratios improve success rate modestly (73\% \(\to\) 93\%) but substantially increase channel usage (10.7 \(\to\) 29.8), reducing explanation compactness.}
\label{fig:sa_ratio_sweep}
\end{figure}

We examine how the group selection ratio \(r\) shapes the trade-off between validity and sparsity in SA counterfactuals. As shown in Figure~\ref{fig:sa_ratio_sweep}, increasing \(r\) from 0.3 to 0.9 raises the counterfactual success rate from 73\% to 93.3\%, but reduces sparsity: the average number of altered channels increases from 10.7 to 29.8 (out of 48), and modified modality groups from 3.9 to 12.7 (out of 16). This finding suggests that static SHAP-based preselection alone cannot achieve both high validity and strong group-level sparsity. This motivates learnable group selection mechanisms that can dynamically prune irrelevant groups during optimization.

\subsection{Learnable Gate Methods vs.\ M-CELS}

To overcome the sparsity limitations of static Shapley-Adaptive methods, we introduce Learnable Gate (LG) variants that jointly optimize trainable group gates with perturbation masks. Table~\ref{tab:lg_methods_vs_mcels} summarizes a comprehensive comparison between LG variants and the channel-level M-CELS baseline across validity, sparsity, proximity, and efficiency metrics. The best-performing variant, LG-SHAP pruned, achieves the highest counterfactual success rate (94.7\%) with strong target confidence (80.8\%), improving upon M-CELS (90.0\% success, 88.2\% confidence). This 4.7 percentage point gain in success rate demonstrates that SHAP-initialized learnable group gating enhances counterfactual reliability while enforcing structured, domain-aligned sparsity. Across sparsity dimensions, LG variants substantially reduce explanation complexity relative to M-CELS. LG-SHAP pruned modifies only 8.2 modality groups and 18.0 channels on average, corresponding to 27\% and 22\% reductions, respectively, while maintaining the highest validity. LG-fixed prune achieves the sparsest explanations (5.2 groups, 11.5 channels) at a modest cost in validity (88.7\%), and LG-scratch yields intermediate performance. All LG variants also converge efficiently, with generation times comparable to or faster than M-CELS (8.0--9.7~s vs.\ 10.0~s). Overall, Table~\ref{tab:lg_methods_vs_mcels} shows that SHAP-initialized learnable gates effectively balance validity, sparsity, and efficiency, producing more interpretable counterfactuals than channel-level baselines.

\input{tables/table1_lg_methods_vs_mcels}

\subsection{Group Sparsity and Plausibility}

\begin{figure}[t]
\centering
\includegraphics[width=0.75\linewidth]{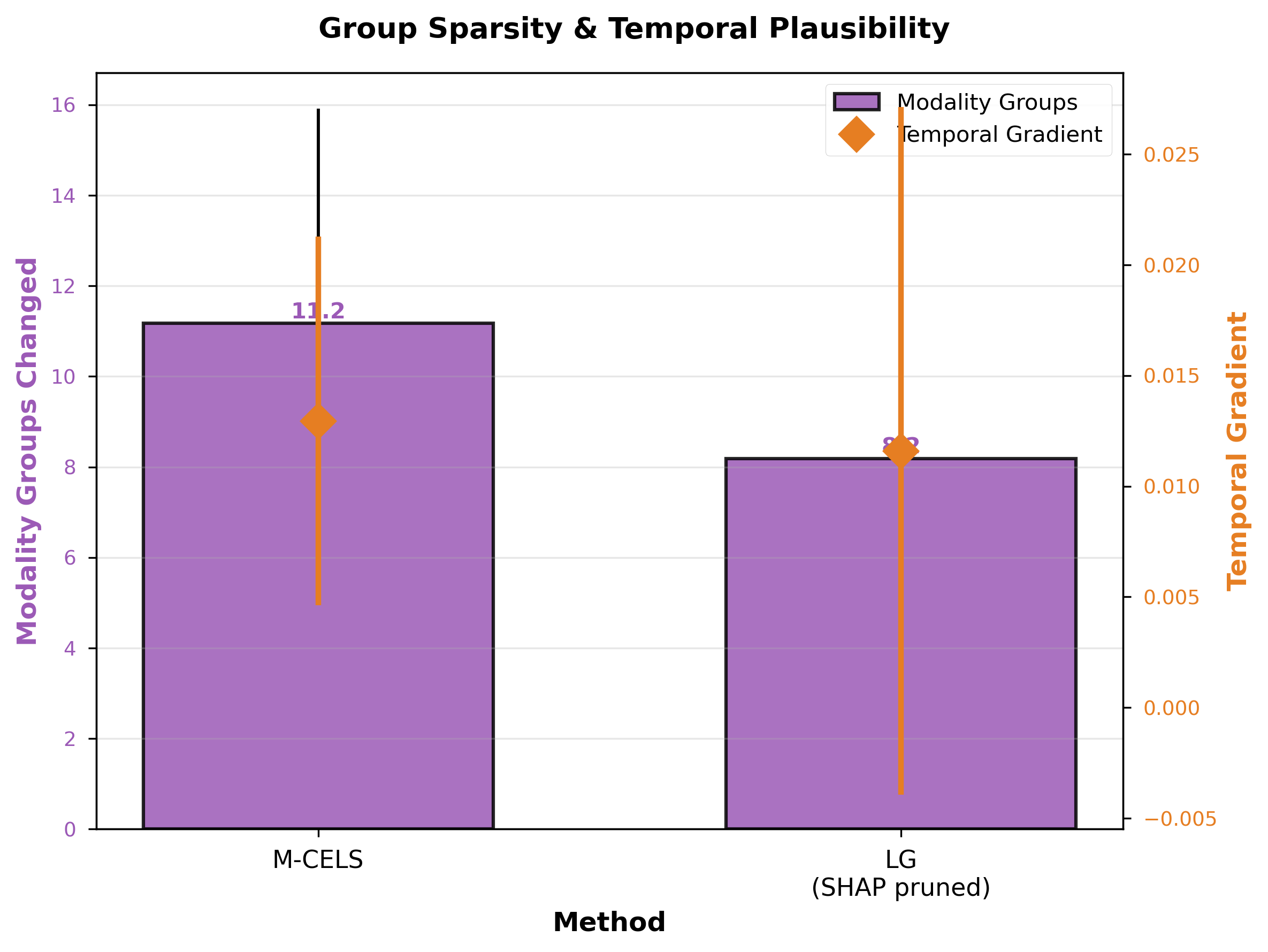}
\caption{Group sparsity and temporal plausibility comparison (\(n = 150\)). LG-SHAP pruned modifies 27\% fewer modality groups (8.2 vs.\ 11.2) while maintaining comparable temporal gradient, demonstrating that structured group-level optimization improves interpretability without sacrificing plausibility.}
\label{fig:lg_mcels_sparsity}
\end{figure}

Beyond validity, interpretable counterfactuals must exhibit both structural sparsity (few modified sensor groups) and temporal plausibility (smooth, realistic modifications). Figure~\ref{fig:lg_mcels_sparsity} compares LG-SHAP pruned and M-CELS on modality-group sparsity and temporal gradient. LG-SHAP pruned achieves 8.2 modified modality groups compared to M-CELS's 11.2 groups, a 27\% reduction that translates to more compact, interpretable explanations. Temporal gradient values indicate comparable temporal smoothness, suggesting that group-structured perturbations do not introduce artificial discontinuities or biomechanically implausible modifications. The dual advantage of stronger structural sparsity with preserved temporal coherence demonstrates that SHAP-initialized learnable group gates successfully encode domain structure into the counterfactual search space while improving validity.

\subsection{Exercise-Specific Correction Patterns}

\begin{figure}[!b]
\centering
\includegraphics[width=\linewidth]{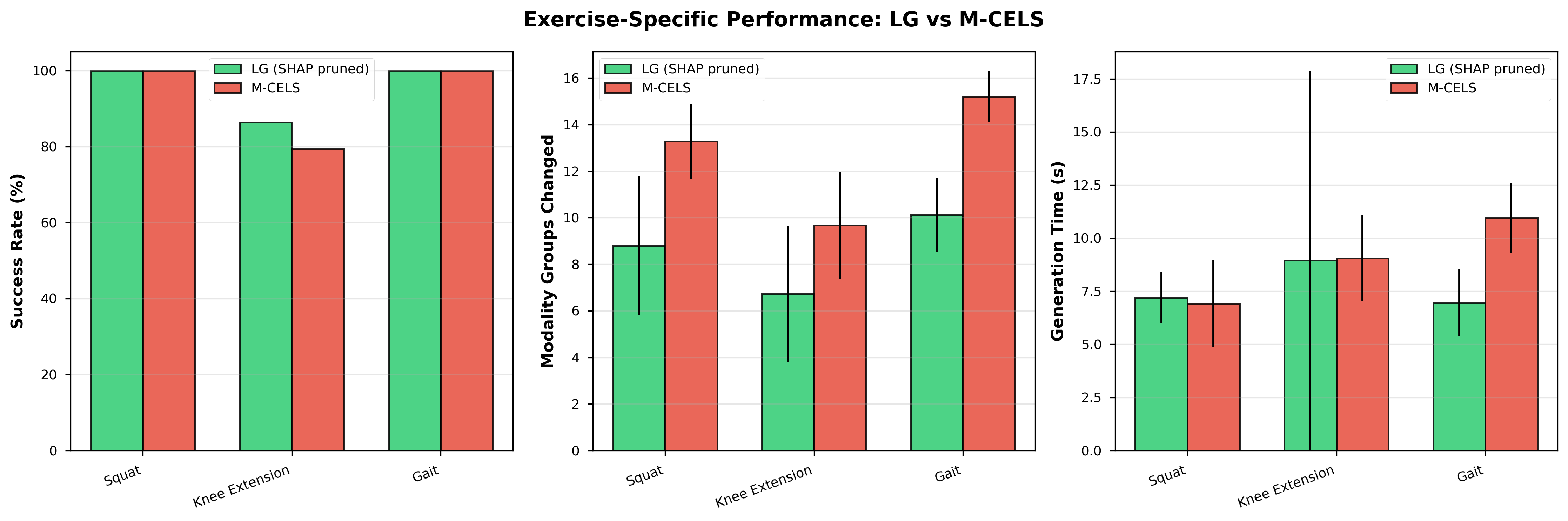}
\caption{Exercise-specific performance comparison between LG-SHAP pruned and M-CELS (\(n = 150\) error-to-correct samples). LG achieves comparable or higher success rates with consistently sparser explanations across all exercise.}
\label{fig:exercise_comparison}
\end{figure}

To assess exercise-specific interpretability, we compare LG (SHAP pruned) and M-CELS across squat, knee extension, and gait tasks. Figure~\ref{fig:exercise_comparison} summarizes counterfactual success, group sparsity, and generation time by exercise. LG achieves comparable or higher success than M-CELS across all exercises, with the largest relative improvement in knee extension, a biomechanically challenging task that requires precise quadriceps activation with minimal hamstring co-contraction. This highlights the advantage of structured group selection for unilateral, range-of-motion–limited movements. Across all exercises, LG consistently produces sparser explanations, modifying fewer modality groups than M-CELS while maintaining validity. Gait counterfactuals involve the largest number of groups, reflecting the multi-joint coordination required for cyclic locomotion, whereas knee extension relies on the fewest groups, consistent with its localized joint mechanics. LG also shows comparable or faster generation times, indicating efficient convergence under group-level constraints.

\begin{figure*}[!ht]
\centering
\includegraphics[width=0.95\linewidth]{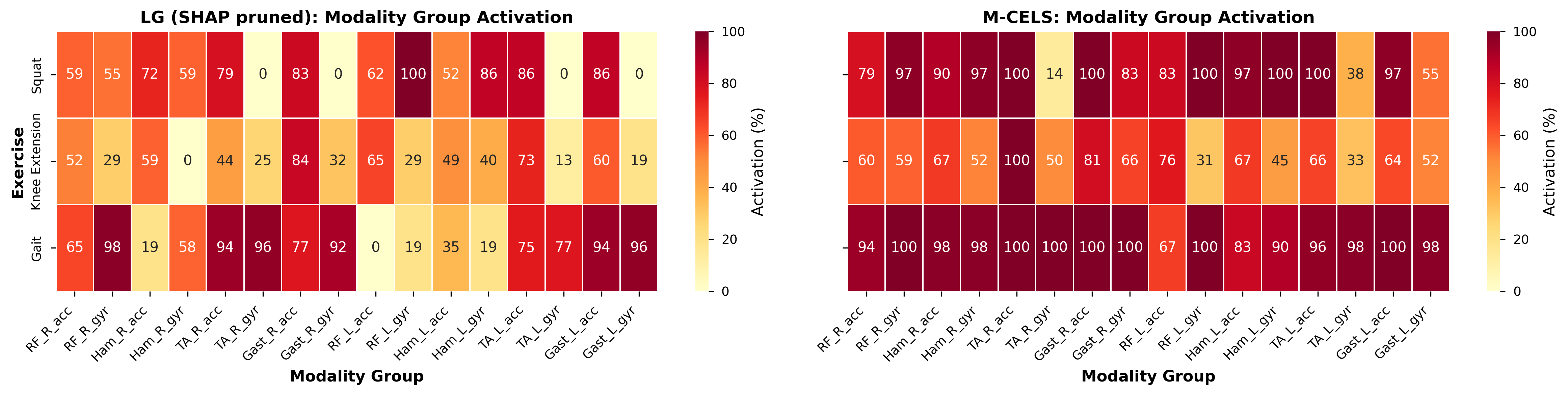}
\caption{Modality group activation frequency for LG (SHAP pruned) vs. M-CELS across exercises. Heatmaps show the proportion of counterfactuals modifying each modality group. LG yields sparser, exercise-specific patterns consistent with biomechanics, whereas M-CELS shows diffuse group activation.}
\label{fig:modality_activation}
\end{figure*}

Figure~\ref{fig:modality_activation} compares modality-group activation frequencies for LG and M-CELS across exercises. LG shows exercise-specific, selective activation, whereas M-CELS activates most groups broadly. In squats, LG emphasizes contralateral limb rotational control to correct weight-transfer asymmetries. In knee extension, LG targets gastrocnemius- and quadriceps-related groups while suppressing hamstring rotations, consistent with reducing antagonistic co-contraction for full extension. In gait, LG focuses on stance-phase rotational dynamics and ankle control while avoiding non-essential groups. Thus, LG achieves structured, exercise-specific sparsity by activating only biomechanically relevant groups, in contrast to M-CELS’s dense, channel-level patterns. These activations map directly to muscle-level clinical guidance: squats prioritize rotational symmetry and bilateral knee control, knee extension emphasizes quadriceps engagement with reduced hamstring co-contraction and ankle stabilization, and gait focuses on stance–swing coordination and push-off mechanics. This group-level sparsity yields concise, anatomically coherent guidance that better supports actionable neuromuscular re-education than diffused channel-level explanations.

\section{Discussion}
\label{sec:discussion}

Domain-aligned sensor grouping improves the interpretability and structure of CEs for high-dimensional IMU time series. Operating at the group level, Adaptive-MO yields explanations consistent with clinicians’ segment- and joint-based reasoning, supporting translation to actionable feedback\cite{Routhier2020,Porciuncula2018}. Improvements in group sparsity, temporal smoothness, and proximity are achieved without sacrificing validity or plausibility; comparable \(L_2\) distances and higher channel sparsity suggest more efficient (not merely redistributed) perturbations. Plausibility scores indicate counterfactuals remain near the data manifold, aligning with counterfactual XAI desiderata\cite{verma2024}. LG-SHAP pruned matches M-CELS generation time while improving validity and sparsity but requires extra setup for Shapley computation and gate initialization; this overhead could be reduced via cached attributions, warm starts, or amortized inference\cite{lundberg2017unified}.

The exercise-specific analysis reveals meaningful patterns showing how group-based counterfactuals support targeted rehabilitation. LG’s improvement over M-CELS on knee extension indicates that structured group-level optimization better respects anatomical constraints by avoiding scattered channel-level changes. Modality-group activations translate into actionable neuromuscular guidance: squat corrections emphasize contralateral limb rotational control to address weight-transfer asymmetries~\cite{Porciuncula2018}; knee extension corrections highlight reduced hamstring co-contraction as key to achieving full extension; and gait corrections prioritize rectus femoris and ankle groups to address stance–swing asymmetries and improve clearance and push-off~\cite{Routhier2020,Porciuncula2018}. Consistently higher group sparsity across exercises further indicates that LG yields fewer simultaneous corrective targets, supporting staged, focused rehabilitation strategies aligned with motor learning principles~\cite{Porciuncula2018}. A key limitation of this work is that experiments are conducted only on the KneE-PAD dataset, which contains a limited number of patient recordings. Evaluating the framework on larger and more diverse rehabilitation datasets would strengthen generalizability. In addition, the sensor grouping is hand-crafted and specific to lower-limb IMUs; extending the approach to other joints or multimodal settings (e.g., IMU+sEMG) will require more flexible grouping strategies that remain guided by biomechanical priors~\cite{zhu2025imu}. Finally, evaluation is currently limited to offline metrics, and future work should include user studies with physiotherapists and patients to assess interpretability, actionability, and trust, as well as longitudinal studies examining rehabilitation outcomes.

\section{Conclusion}
\label{sec:conclusion}

This paper introduces an adaptive multi-objective framework for group-based CEs in multivariate time-series data, applied to IMU-based knee rehabilitation. Combining Shapley-based group ranking, learnable group gates, and structured multi-objective optimization, it generates sparse, temporally coherent, and physiologically meaningful counterfactuals. On the KneE-PAD dataset, it markedly improves sparsity, group structure, and temporal smoothness over a strong channel-level baseline (M-CELS) while maintaining or improving validity. The results show that dynamic feature group-level optimization yields more interpretable and reliable explanations than channel-level methods in IMU-based rehabilitation classification. The work also highlights sensor grouping as a theoretically grounded, practically important design choice for time-series counterfactuals. Future directions include automatic group discovery, multimodal fusion with sEMG and clinical covariates, and real-time telerehabilitation deployment.

\section*{Acknowledgments}

This research was funded by the Dutch Research Council (NWO) within the framework of the LoaD project. 
We express our gratitude to the contributors of the KneE-PAD dataset and to all participating patients for their valuable involvement.
\bibliographystyle{ieeetr}
\bibliography{references}
\end{document}

%% file: tables/table1_lg_methods_vs_mcels.tex

\begin{table}[!b]
\centering
\caption{LG methods vs. M-CELS on KneE-PAD ($n=150$). Mean $\pm$ std.}
\label{tab:lg_methods_vs_mcels}
\scriptsize
\setlength{\tabcolsep}{3pt}
\begin{tabular}{lcccccc}
\hline
Method & Succ. & Conf. & Grp. & Chn. & $L_2$ & Time \\
       & (\%)  & (\%)  & \#   & \#  & Dist. & (s) \\
\hline
M-CELS & 90.0$\pm$30.1 & \textbf{88.2}$\pm$29.5 & 11.2$\pm$4.7 & 23.2$\pm$11.6 & 45.8$\pm$19.8 & 10.0$\pm$3.5 \\
LG-SHAP & \textbf{94.7}$\pm$22.5 & 80.8$\pm$22.7 & 8.2$\pm$3.1 & 18.0$\pm$11.0 & 50.8$\pm$19.5 & 9.7$\pm$4.2 \\
LG-scratch & 90.7$\pm$29.2 & 79.3$\pm$27.9 & 10.8$\pm$4.5 & 25.0$\pm$16.4 & 53.7$\pm$22.5 & 9.3$\pm$4.4 \\
LG-fixed & 88.7$\pm$31.8 & 75.2$\pm$28.7 & \textbf{5.2}$\pm$3.2 & \textbf{11.5}$\pm$8.6 & \textbf{43.5}$\pm$20.7 & \textbf{8.0}$\pm$3.9 \\
\hline
\end{tabular}
\end{table}